\documentclass{ecai} 

\usepackage{soul}
\usepackage{url}
\usepackage[utf8]{inputenc}

\usepackage{algorithm}
\usepackage{algorithmic}
\usepackage[switch]{lineno}
\usepackage{tcolorbox}

\usepackage{latexsym}
\usepackage{amssymb}
\usepackage{amsmath}
\usepackage{amsthm}
\usepackage{booktabs}
\usepackage{enumitem}
\usepackage{graphicx}
\usepackage{color}
\usepackage{multirow}
\usepackage{tikz}
\usepackage{xspace}

\newcommand\anonym[1]{}

\newcommand{\orangesum}{OrangeSum\xspace}

\newcommand{\falc}{ETR\xspace}

\newcommand{\barthez}{BARThez\xspace}

\newcommand{\barthezosum}{$\text{BARThez}^{\ast}$\xspace}

\newcommand\nextToken\relax

\begin{document}

\begin{frontmatter}
\paperid{0534}
\title{Inclusive Easy-to-Read Text Generation for Individuals with Cognitive Impairments}

\author[A,C]{\fnms{François}~\snm{Ledoyen}\thanks{Corresponding Author. Email: ledoyenfrancois@gmail.com}}
\author[A]{\fnms{Gaël}~\snm{Dias}}
\author[A]{\fnms{Alexis}~\snm{Lechervy}}
\author[A]{\fnms{Jérémie}~\snm{Pantin}}
\author[A]{\fnms{Fabrice}~\snm{Maurel}}
\author[A]{\fnms{Youssef}~\snm{Chahir}}
\author[B]{\fnms{Elisa}~\snm{Gouzonnat}}
\author[B]{\fnms{Mélanie}~\snm{Berthelot}}
\author[B]{\fnms{Stanislas}~\snm{Moravac}}
\author[C]{\fnms{Armony}~\snm{Altinier}}
\author[C]{\fnms{Amy}~\snm{Khairalla}}

\address[A]{Université Caen Normandie, ENSICAEN, CNRS, Normandie Univ, GREYC UMR 6072, F-14000 Caen, France}
\address[B]{Université Caen Normandie, CRISCO UR 4255, F-14000 Caen, France}
\address[C]{Koena SAS, F-31450 Fourquevaux, France}

\begin{abstract}
Ensuring accessibility for individuals with cognitive impairments is essential for autonomy, self-determination, and full citizenship. 
However, manual Easy-to-Read (ETR) text adaptations are slow, costly, and difficult to scale, limiting access to crucial information in healthcare, education, and civic life. 
AI-driven ETR generation offers a scalable solution but faces key challenges, including dataset scarcity, domain adaptation, and balancing lightweight learning of Large Language Models (LLMs). 
In this paper, we introduce ETR-fr, the first dataset for ETR text generation fully compliant with European ETR guidelines. 
We implement parameter-efficient fine-tuning on PLMs and LLMs to establish generative baselines. 
To ensure high-quality and accessible outputs, we introduce an evaluation framework based on automatic metrics supplemented by human assessments. 
The latter is conducted using a 36-question evaluation form that is aligned with the guidelines.
Overall results show that PLMs perform comparably to LLMs and adapt effectively to out-of-domain texts.
Code and datasets are available at \url{https://github.com/FrLdy/ETR-fr}.
\end{abstract}

\end{frontmatter}

\section{Introduction}
\begin{figure}[!ht]
    \centering
    \begin{minipage}{0.5\columnwidth}
        \fbox{\includegraphics[width=.95\textwidth,page=1]{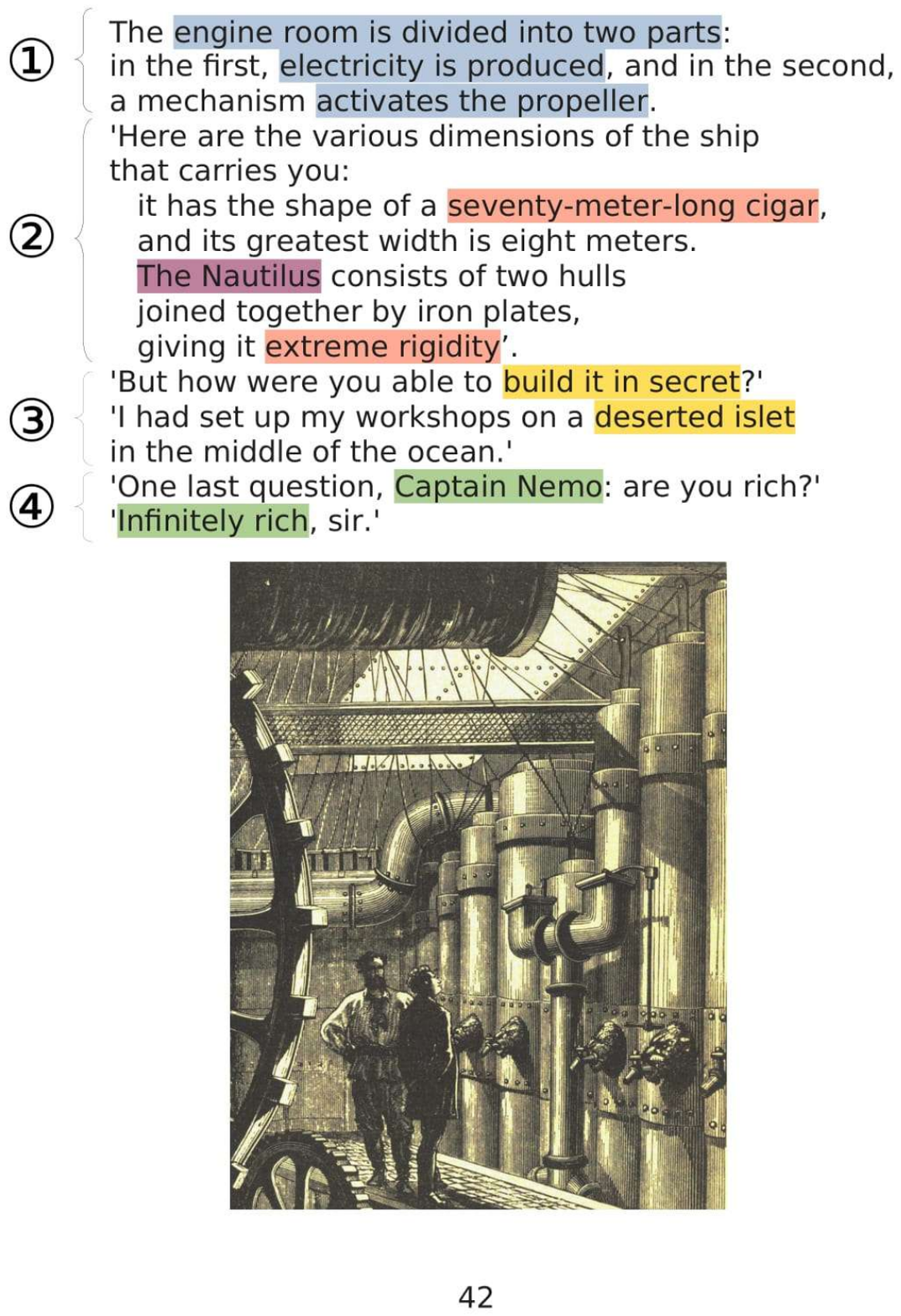}}
    \end{minipage}%
    \begin{minipage}{0.5\columnwidth}
        \fbox{\includegraphics[width=.95\linewidth,page=2]{figures/nautilus.pdf}}
    \end{minipage}
    
    \caption{Extract of the Easy-to-Read book \textit{Twenty Thousand Leagues Under the Sea} by Jules Verne from François~Baudez Publishing. 
    The original document is in French, but we translated it into English to ease comprehension.
    \textbf{Left page} is the original text with an illustration. 
    \textbf{Right page} is the ETR transcription with the main information plus its captioned {\it vignettes}.
    We have highlighted and numbered the paragraphs to show the matches between the original and the ETR versions.
    }
    \label{fig:example_etr}
\end{figure}

Reflecting the priorities of global initiatives such as the United Nations Sustainable Development Goals\footnote{\url{https://sdgs.un.org/goals}} and the Leave No One Behind Principle\footnote{\url{https://unsdg.un.org/2030-agenda/universal-values/leave-no-one-behind}}, ensuring accessibility for individuals with cognitive impairments is crucial to fostering autonomy, self-determination, and full citizenship. Individuals with intellectual disabilities deserve equal rights to participate in society, to make informed choices, and to fully engage in their communities. However, they continue to face significant obstacles, especially in accessing written information, which is essential for healthcare, education, employment, and civic engagement.
Mental health disorders and intellectual disabilities affect millions worldwide, with an estimated 1.3\% of the global population experiencing significant cognitive challenges \cite{MAULIK2011419}. In Europe alone, 4.2 million individuals are affected, while in France, between 650,000 and 700,000 people live with intellectual disabilities that limit their ability to comprehend written materials \cite{GUSTAVSSON2011718}. 

Easy-to-Read (ETR) is a well-established method for simplifying complex documents,
ensuring that people with cognitive impairments can understand and use key information autonomously~\cite{pathways_information_2021}. European organizations and institutions, including France's National Solidarity Fund for Autonomy\footnote{\url{https://www.cnsa.fr/}}, are increasingly producing simplified materials, indicating growing recognition of its value in improving accessibility for diverse populations. However, the current manual adaptation process is slow, costly, and subject to strict certification requirements, making it difficult to scale \cite{chebab_falc_2019}.

Developing effective AI-driven accessibility tools comes with several challenges. One major obstacle is the construction of high-quality datasets, ensuring that AI models learn to generate clear and meaningful adapted texts. Additionally, a balance must be struck between parameter-efficient fine-tuning (PEFT) approaches, which enable low-resource, efficient adaptation, and large language model (LLM) based techniques, which leverage extensive linguistic knowledge for high-quality text simplification.
Open-source development ensures transparency and collaboration while empowering individuals to customize solutions and fully participate as equal citizens.

Generating high-quality ETR texts is challenging due to the need for linguistic simplification and strict adherence to accessibility guidelines. To address these challenges, we introduce ETR-fr, the first dataset specifically designed for ETR text generation, tailored to users with cognitive disabilities. This dataset comprises 523 aligned text pairs and fully complies with European ETR guidelines. We develop generative models using PEFT strategies, such as prefix-tuning \cite{li_prefix-tuning_2021} and Low-Rank Adaptation (LoRA) \cite{hu2022lora} applied to pretrained language models like mBART \cite{liu_multilingual_2020} and mBARThez \cite{kamal_eddine_barthez_2021}, as well as large language models like Mistral-7B \cite{jiang_mistral_2023} and Llama-2-7B \cite{touvron_llama_2023}. On the other hand, to ensure the highest quality in generating accessible texts, rigorous evaluation is essential. The different generative models undergo intrinsic evaluation using a comprehensive set of automatic metrics derived from text simplification and text summarization. However, given the critical need for clarity, coherence, and accessibility in this context, manual evaluation plays a central role.
Our main contributions are summarized as follows:
\begin{itemize}
    \item Introduction of ETR-fr, the first parallel dataset fully compliant with European ETR guidelines.
    \item Implementation of baselines for ETR generation based on PEFT strategies, such as prefix-tuning and LoRA, applied to PLMs and LLMs backbones.
    \item Comprehensive evaluation framework using intrinsic metrics from text simplification and summarization, reinforced by a 36-question manual assessment based on European ETR guidelines.
    \item Investigation of the model's ability to generalize ETR generation from our ETR-fr to politically focused materials.
\end{itemize}
\section{Easy-to-Read Framework} \label{sec:falc_guidelines}

Creating accessible texts for individuals with cognitive disabilities follows the Easy-to-Read framework, which adapts content to align with the European Easy-to-Read guidelines \cite{pathways_information_2021} (see example in Figure~\ref{fig:example_etr}). The key principles are outlined as follows.

\paragraph{Clear and simple language:} Use everyday vocabulary, avoiding technical jargon. Sentences should be short, direct, and in the active voice to specify who is performing an action. Each sentence should convey only one idea, and consistent terminology should be used throughout the text.

\paragraph{{Examples and analogies:}} Provide concrete examples and relatable analogies to explain abstract or complex ideas, linking them to familiar situations for better comprehension.

\paragraph{{Structure and organization:}} Arrange content into clearly defined sections with descriptive headings and subheadings. Information should follow a logical sequence, grouping related concepts and using bullet lists while avoiding lengthy paragraphs 

\paragraph{{Accessible content:}} Begin with a summary outlining key points in simple terms. If technical terms are necessary, introduce clear definitions. For complex concepts or procedures, explain each step systematically with concrete examples.  

\paragraph{{Visuals and illustrations:}} Incorporate relevant images, charts, or diagrams to reinforce key messages. Visuals should be simple, directly connected to the text, and include concise explanatory captions.

Following the ETR guidelines, ensuring the validity of ETR content requires approval from both experts and the target audience. The manual ETR transcription process involves summarizing content and simplifying it through an iterative collaboration between human experts and individuals with cognitive impairments. This co-creation process is essential for obtaining the official European ETR label \cite{pathways_information_2021}.
\section{Related Work}

Automating ETR generation could significantly streamline document creation and bridge the digital divide. However, research in this area remains scarce, except for very few studies mainly conducted in Europe \cite{CallejaEM24,martinez_exploring_2024}. In contrast, related fields such as text simplification \cite{AsthanaRCHL24, li_investigating_2024} and text summarization \cite{zakkas_sumblogger_2024} have been widely studied.

Within the natural language processing field, various studies and tools have been developed to support individuals with cognitive disabilities by enhancing augmentative communication methods \cite{martin_bridging_2024,murillo-morales_automatic_2020}, with dialogue agents being a widely explored solution \cite{huq_dialogue_2024}.

\begin{table*}[!ht]
\caption{
Statistics between ETR-fr, OrangeSum, Alector, Finnish-Easy, and ClearSim datasets. 
Compression and novelty ratios are not given for ClearSim as the publicly available version is not aligned. The LIX readability index is used instead of KMRE for Finnish-Easy and ClearSim, as it is language-independent. Results are given on average with corresponding standard deviation over documents.
}

\label{tab:dataset_comparaison}
\centering
\footnotesize
\renewcommand{\arraystretch}{1.4}
\begin{tabular*}{\textwidth}{l@{\extracolsep{\fill}}lccccc@{\extracolsep{\fill}}}
\toprule
                           &                 & \multicolumn{3}{c}{\textbf{French}}                             & \multicolumn{2}{c}{\textbf{Finnish and Spanish}}             \\
\cmidrule(lr){3-5} \cmidrule(l){6-7}
                           &                 & \textbf{ETR-fr (ours)} & \textbf{Alector} & \textbf{OrangeSum} & \textbf{Finnish-Easy} & \textbf{ClearSim} \\
\midrule
\textbf{Dataset size}      &                 & 523                    & 79               & 24,401              & 1587                   & 207               \\
\midrule
\multirow{2}{*}{\textbf{Vocabulary size}}   & \textbf{source} & 4547                   & 3129             & 80,295              & 98,833                 & 6067              \\
                           & \textbf{target} & 1765                   & 2538             & 23,092              & 18,934                 & 2952              \\
\midrule
\multirow{2}{*}{\textbf{Num. of words}}     
                           & \textbf{source} & $102.76_{\pm 42.84}$   & $306.48_{\pm 90.83}$ & $375.98_{\pm 183.34}$ & $348.47_{\pm 266.71}$ & $429.13_{\pm 225.28}$ \\
                           & \textbf{target} & $46.15_{\pm 16.73}$    & $285.63_{\pm 85.34}$ & $34.00_{\pm 12.17}$   & $55.00_{\pm 16.61}$   & $147.78_{\pm 59.54}$ \\
\midrule
\multirow{2}{*}{\textbf{Num. of sentences}} 
                           & \textbf{source} & $9.30_{\pm 5.12}$      & $20.56_{\pm 8.95}$   & $17.15_{\pm 8.85}$    & $30.82_{\pm 24.05}$   & $23.00_{\pm 12.77}$ \\
                           & \textbf{target} & $7.13_{\pm 3.85}$      & $22.72_{\pm 9.79}$   & $1.86_{\pm 0.94}$     & $6.97_{\pm 2.13}$     & $11.88_{\pm 5.44}$  \\
\midrule
\multirow{2}{*}{\textbf{Sentence length}}    
                           & \textbf{source} & $12.57_{\pm 5.63}$     & $16.82_{\pm 6.14}$   & $22.77_{\pm 5.99}$    & $11.29_{\pm 1.83}$    & $20.13_{\pm 9.21}$  \\
                           & \textbf{target} & $7.89_{\pm 4.55}$      & $13.87_{\pm 4.08}$   & $21.68_{\pm 10.82}$   & $8.04_{\pm 1.55}$     & $13.04_{\pm 6.61}$  \\
\midrule
\multirow{2}{*}{\textbf{KMRE $\uparrow$}}    
                           & \textbf{source} & $91.43_{\pm 9.41}$     & $88.56_{\pm 8.23}$   & $69.80_{\pm 9.47}$    & –                     & –                  \\
                           & \textbf{target} & $98.94_{\pm 10.60}$    & $95.25_{\pm 7.15}$   & $68.32_{\pm 16.07}$   & –                     & –                  \\
\midrule
\multirow{2}{*}{\textbf{LIX $\downarrow$}}   
                           & \textbf{source} & $33.59_{\pm 8.72}$     & $39.06_{\pm 9.44}$   & $49.95_{\pm 7.90}$    & $67.44_{\pm 5.82}$    & $59.12_{\pm 8.89}$  \\
                           & \textbf{target} & $26.89_{\pm 9.68}$     & $34.19_{\pm 8.27}$   & $50.39_{\pm 13.43}$   & $58.12_{\pm 8.47}$    & $45.30_{\pm 10.24}$ \\
\midrule
\textbf{Comp. ratio (\%)}  &                 & $50.05_{\pm 20.55}$    & $6.84_{\pm 4.47}$    & $89.16_{\pm 6.34}$    & $75.40_{\pm 21.71}$   & –                  \\
\textbf{Novelty (\%)}      &                 & $53.80_{\pm 16.14}$    & $17.84_{\pm 8.72}$   & $38.24_{\pm 19.71}$   & $54.74_{\pm 16.55}$   & –                  \\
\bottomrule
\end{tabular*}

\end{table*}

Within the context of inclusive text generation, \citet{goodman_lampost_2022} introduced an email-writing interface based on LaMDA LLM \cite{thoppilan_lamda_2022},
offering features such as summarization, subject line generation, and text revision. 
However, human evaluations show that current LLMs still lack accuracy and quality for dyslexic users, highlighting the need for further research.
In French, the Hector system \cite{todirascu_hector_2022} integrates word embeddings with rule-based methods for adapting text to be dyslexia-friendly. While syntactic transformations improve readability, results show a decline in performance at the discourse and lexical levels.

Within the specific domain of ETR generation, \citet{dmitrieva_towards_2024} have created the Finnish-Easy dataset, which aligns news articles with their Easy Finnish\footnote{Easy Finnish is a form of Finnish where the language has been adapted so that it is easier to read and understand in terms of content, vocabulary and structure.} (\textit{selkosuomi}) versions through automatic alignment. However, the authors acknowledge potential inaccuracies in text pairing and note that Easy Finnish does not strictly adhere to ETR guidelines. Additionally, they introduce baseline models for ETR sentence generation using fine-tuned mBART and FinGPT \cite{luukkonen_fingpt_2023}. 
Similarly, the ClearText project \cite{espinosa-zaragoza_automatic_2023} aims to develop the ClearSim corpus for simplifying Spanish public administrative texts. The current public version\footnote{\url{https://github.com/gplsi/corpus-cleartext-cas-v1.0/tree/main}} contains three ETR document pairs with 201 misaligned pages, limiting its suitability for learning purposes. However, the project plans to expand the corpora to 18,000 texts, 15,000 generated by ChatGPT and 3,000 transcribed by experts. 
More recently, \citet{martinez_exploring_2024} introduced an automatically aligned Spanish ETR corpus alongside a fine-tuned Llama-2-7B model. An expert-led evaluation highlights progress in accessibility and underscores ongoing challenges in producing high-quality, guideline-compliant document-level generation. This study highlights the challenges of cross-lingual transfer, demonstrating that the translate-simplify-retranslate strategy often leads to incorrect or untranslated outputs.

Although these initiatives reflect a growing interest in ETR generation, they highlight the absence of high-quality resources that fully adhere to the European ETR guidelines. To address this gap, we introduce ETR-fr, the first expert-transcribed ETR dataset specifically designed for users with cognitive disabilities.

\section{ETR-fr Dataset}

Although several datasets exist for French text simplification and summarization, such as Alector \cite{gala_alector_2020}, OrangeSum \cite{kamal_eddine_barthez_2021}, and multilingual corpora \cite{hauser-etal-2022-multilingual}, there remains a lack of high-quality, document-aligned corpora specifically designed for ETR text generation. This gap is particularly noticeable for the French language.

To address this, we introduce the {ETR-fr dataset}, constructed from the \textit{Facile à Lire et à Comprendre} (Easy-to-Read-and-Understand)\footnote{Known in French as \textit{Facile à Lire et à Comprendre}.} collection published by François Baudez Publishing\footnote{\url{http://www.yvelinedition.fr/Facile-a-lire}}. This collection consists of eleven children's books adapted according to European guidelines for cognitive accessibility. Each book presents the original version on the left page and its ETR transcription on the right, as illustrated in Figure~\ref{fig:example_etr}.

From these books, we extracted 523 aligned page pairs $(\text{source}, \text{target})$, where the source corresponds to the original text and the target to its ETR version. These alignments form the core of the ETR-fr dataset.

\paragraph{ETR-fr Characteristics}

Table~\ref{tab:dataset_comparaison} summarizes the characteristics of the dataset, including readability metrics, compression ratios, and novelty rates. 
We use two readability indicators: {KMRE}~\cite{kandel_application_1958}, a French adaptation of the Flesch-Kincaid Reading Ease formula~\cite{kincaid_derivation_1975}, and {LIX}~\cite{bjornsson_readability_1983}. KMRE produces a score from 0 (very difficult) to 100 or more (very easy), based on sentence and word lengths. LIX measures difficulty based on average sentence length and the proportion of long words (more than six letters), with typical values ranging from 20 (easy) to 60 (difficult).
Novelty \cite{narayan_dont_2018} indicates the proportion of new unigrams introduced in the target text.

On average, ETR-fr achieves a {50.05\% compression rate}, reducing token count by 56.61 and sentence count by 2.17. 
The average novelty rate is 53.80\%. 
The KMRE score improves by 7.51 points on average, showing a measurable gain in readability from source to ETR output.

\paragraph{Comparison with Related Datasets}

To better contextualize ETR-fr, we compare it with other French-language datasets: {Alector} \cite{gala_alector_2020}, designed for text simplification, and {OrangeSum} \cite{kamal_eddine_barthez_2021}, built for summarization. As shown in Table~\ref{tab:dataset_comparaison}, OrangeSum features a high compression rate (89.16\%) but reduced readability in its target texts. Alector presents minimal compression (6.84\%) but improves readability by 6.69 KMRE points. ETR-fr offers a more balanced profile, combining moderate compression (50.05\%), a readability improvement of 7.51 KMRE points, and higher novelty (53.80\%) than both OrangeSum (38.24\%) and Alector (17.84\%).

We also compare ETR-fr to foreign-language ETR-style datasets: {Easy-Finnish} \cite{dmitrieva_towards_2024} and {ClearSim} \cite{espinosa-zaragoza_automatic_2023}. These corpora are designed for broader audiences and focus on various text types. Easy-Finnish covers news articles, and ClearSim includes administrative texts. Easy-Finnish demonstrates a high compression rate (75.40\%) and a novelty score similar to ETR-fr ($\simeq$ 54\%). However, both datasets exhibit lower accessibility, with LIX readability scores significantly higher than those of ETR-fr: +33.85 and +25.53 for source texts, and +31.23 and +18.41 for target texts. Additionally, ClearSim does not include reliable compression and novelty statistics due to misalignment in its text pairs.

\paragraph{ETR-fr Splits}
\begin{table*}[!ht]
\caption{Statistics for the ETR-fr dataset (Train/Validation/Test) and the ETR-fr-politic test set. Results are given on average with corresponding standard deviation over documents.}
\label{tab:dataset_splits}
\centering
\footnotesize
\resizebox{\linewidth}{!}{
\renewcommand{\arraystretch}{1.4}
\begin{tabular}{lcccccccc}
\toprule
\textbf{} & \multicolumn{6}{c}{\textbf{ETR-fr}} & \multicolumn{2}{c}{\textbf{ETR-fr-politic}} \\
\cmidrule(lr){2-7} \cmidrule(l){8-9}

\textbf{} & \multicolumn{2}{c}{\textbf{Train}} & \multicolumn{2}{c}{\textbf{Validation}} & \multicolumn{2}{c}{\textbf{Test}} & \multicolumn{2}{c}{\textbf{Test}} \\
\cmidrule(lr){2-3} \cmidrule(lr){4-5} \cmidrule(lr){6-7} \cmidrule(l){8-9}
& \textbf{source} & \textbf{target} & \textbf{source} & \textbf{target} & \textbf{source} & \textbf{target} & \textbf{source} & \textbf{target} \\
\midrule
\textbf{Num. of texts} & \multicolumn{2}{c}{$399$} & \multicolumn{2}{c}{$71$} & \multicolumn{2}{c}{$53$} & \multicolumn{2}{c}{$33$} \\
\textbf{Num. of words} & $99.70_{\pm 39.25}$ & $46.50_{\pm 16.80}$ & $100.76_{\pm 48.12}$ & $48.59_{\pm 17.20}$ & $128.47_{\pm 52.54}$ & $40.26_{\pm 14.38}$ & $96.27_{\pm 56.34}$ & $62.85_{\pm 30.04}$ \\
\textbf{Num. of sentences} & $8.92_{\pm 4.73}$ & $7.48_{\pm 3.42}$ & $9.03_{\pm 5.21}$ & $7.77_{\pm 3.91}$ & $12.51_{\pm 6.60}$ & $10.34_{\pm 3.81}$ & $6.42_{\pm 3.17}$ & $6.09_{\pm 2.87}$ \\
\textbf{Sentence length} & $12.57_{\pm 4.53}$ & $6.92_{\pm 2.91}$ & $13.59_{\pm 10.53}$ & $6.90_{\pm 2.30}$ & $11.16_{\pm 2.86}$ & $3.97_{\pm 0.88}$ & $15.68_{\pm 6.32}$ & $11.47_{\pm 7.21}$ \\
\textbf{KMRE $\uparrow$} & $91.03_{\pm 8.67}$ & $99.71_{\pm 9.43}$ & $89.50_{\pm 13.49}$ & $100.59_{\pm 10.30}$ & $97.02_{\pm 5.48}$ & $103.67_{\pm 10.71}$ & $75.03_{\pm 11.15}$ & $88.12_{\pm 11.34}$ \\
\textbf{Compression (\%)} & \multicolumn{2}{c}{$49.04_{\pm 20.12}$} & \multicolumn{2}{c}{$44.47_{\pm 22.10}$} & \multicolumn{2}{c}{$65.19_{\pm 14.18}$} & \multicolumn{2}{c}{$29.17_{\pm 22.48}$} \\
\textbf{Novelty (\%)} & \multicolumn{2}{c}{$53.79_{\pm 16.32}$} & \multicolumn{2}{c}{$52.96_{\pm 16.24}$} & \multicolumn{2}{c}{$55.01_{\pm 14.80}$} & \multicolumn{2}{c}{$63.78_{\pm 13.85}$} \\
\bottomrule
\end{tabular}
}
\end{table*}

The ETR-fr dataset is divided into training, validation, and test sets, as described in Table~\ref{tab:dataset_splits}. Two books are selected for the test set to maximize diversity in sentence structure, length, compression, novelty, and readability. The remaining nine books are split into training and validation subsets using a stratified approach.

\paragraph{ETR-fr-politic Test Set}

The participation of persons with disabilities in political and public life is enshrined in the United Nations Convention on the Rights of Persons with Disabilities, which France has ratified. Since 2021, candidates for the French presidential election have been required to submit an ETR version of their electoral programs. 

To assess the robustness of ETR models and their ability to generalize across diverse and previously unseen domains, we evaluate them on a test set specifically focused on political election texts. It is important to note that political texts were not part of the training data, making this evaluation a critical measure of model generalization.

To this end, we introduce an out-of-domain test set, ETR-fr-politic, comprising 33 paragraph pairs manually extracted from the ETR-labeled versions of the 2022 French presidential election programs\footnote{\url{https://www.cnccep.fr/candidats.html}}. These paragraphs have been carefully aligned with their original versions, allowing for precise quantitative evaluation of generated texts. A detailed overview of the dataset is provided in Table~\ref{tab:dataset_splits}.

Compared to the ETR-fr test set, ETR-fr-politic contains fewer texts (33 vs. 53), and its source texts are shorter in both word count (96.27 vs. 128.47) and sentence count (6.42 vs. 12.51). However, its target texts are longer, averaging 62.85 words compared to 40.26 in ETR-fr. The ETR-fr test set exhibits higher readability, with KMRE scores of 97.02 (source) and 103.67 (target), versus 75.03 and 88.12 in ETR-fr-politic. Furthermore, ETR-fr has a higher compression ratio (65.19\%) and a slightly lower novelty rate (55.01\%) compared to ETR-fr-politic (29.17\% and 63.78\%, respectively). These differences highlight the more complex and varied nature of the political texts.

\paragraph{Summary}
In summary, the ETR-fr dataset fills an important gap in French-language NLP resources by providing a high-quality, document-aligned corpus tailored for readers with cognitive impairments. It effectively bridges simplification and summarization, improving readability while maintaining moderate compression and incorporating a high level of novel content. Its structure and evaluation design make it well-suited as a benchmark for training and evaluating ETR generation systems.
\section{ETR Generation and Evaluation}

To evaluate generation models on ETR-fr and establish baseline performance, we design a learning benchmark that involves parameter-efficient fine-tuning of pre-trained language models (PLMs) and LLMs. Our approach also incorporates a two-step pipeline combining text simplification and summarization, mimicking a human-expert strategy.

\subsection{Expert-Centric Configuration}

We introduce an expert-centric pipeline motivated by the lack of established ETR benchmarks and inspired by manual transcription practices. 
This approach replicates the traditional two-step process used by experts, where summarization precedes simplification. Following the methodology proposed by \citet{blinova_simsum_2023}, our pipeline first applies a document-level summarization model, using BARThez trained on OrangeSum \cite{kamal_eddine_barthez_2021}, and then simplifies the output with the MUSS model \cite{martin_muss_2022}, which performs sentence-level simplification using default control tokens. Since neither model is fine-tuned on ETR-fr, this setup allows us to evaluate the zero-shot performance of task-specific models on ETR-fr. 

\subsection{Parameter-Efficient Fine-Tuning}

To conduct ETR generation, we also investigate parameter-efficient fine-tuning (PEFT) of sequence-to-sequence models, which are widely employed in the context of abstractive summarization and text simplification, such as mBART~\cite{liu_multilingual_2020} and mBARThez~\cite{kamal_eddine_barthez_2021}. Additionally, we explore the performance of LLMs, namely Mistral-7B~\cite{jiang_mistral_2023} and Llama-2-7B~\cite{touvron_llama_2023} under PEFT. 

With the growing sophistication of PLMs and LLMs, reducing computational costs while maintaining performance has become a priority. This has led to the development of PEFT strategies, such as prefix-tuning~\cite{li_prefix-tuning_2021} and low-rank adaptation (LoRA)~\cite{hu2022lora}. These methods enable fine-tuning of only a small subset of parameters while keeping most model weights frozen, thereby minimizing the risk of catastrophic forgetting \cite{vu_overcoming_2022}.

\newcommand{\dmodel}{d_{\text{model}}}
\newcommand{\dhead}{d_{\text{head}}}

\begin{table*}[!ht]
\centering
\caption{
Performance of expert-centric and fine-tuned models on the ETR-fr test set (FT: full fine-tuning, PT: prefix-tuning).
\barthezosum denotes \barthez fine-tuned on \orangesum~\cite{kamal_eddine_barthez_2021}.
Scores are averaged over 5 runs (except pipelines), with standard deviation. 
Best results are in bold, except for novelty and compression, where values closest to the ETR-fr test set (Table~\ref{tab:dataset_splits}) are highlighted.}

\label{tab:results}
\footnotesize
\resizebox{\textwidth}{!}{
\renewcommand{\arraystretch}{1.4}
\begin{tabular}{llcccccccc}
\toprule
 &  & { \textbf{ROUGE-1}} & { \textbf{ROUGE-2}} & { \textbf{ROUGE-L}} & { \textbf{BERT-$F_1$}} & { \textbf{SARI}} & { \textbf{KMRE}} & { \textbf{Comp. ratio}} & { \textbf{Novelty}}\\
 \midrule
{\textbf{Expert-centric}} &  &  &  &  &  &  &  &  & \\
 \midrule
\barthezosum &  & $22.85$ & $5.30$ & $15.28$ & $67.54$ & $36.87$ & $95.26$ & $73.38$ & $30.17$\\
MUSS &  & $28.11$ & $8.87$ & $18.54$ & $70.92$ & $36.48$ & $98.03$ & $6.62$ & $15.00$\\
\barthezosum+MUSS &  & $22.42$ & $4.48$ & $14.64$ & $67.58$ & $36.70$ & $96.70$ & $75.61$ & $36.51$\\
MUSS+\barthezosum &  & $20.15$ & $5.36$ & $13.58$ & $66.85$ & $37.56$ & $93.74$ & $75.62$ & $\mathbf{37.48}$\\
 \midrule
{\textbf{Fine-Tuning}} &  &  &  &  &  &  &  &  & \\
 \midrule
\multirow{2}{*}{Mistral-7B} & PT & $23.78_{\pm 12.03}$ & $8.33_{\pm 4.70}$ & $16.90_{\pm 8.20}$ & $64.44_{\pm 15.14}$ & $38.21_{\pm 1.36}$ & $98.99_{\pm 0.80}$ & $30.88_{\pm 18.92}$ & $6.20_{\pm 5.18}$\\
 & LoRA & $30.53_{\pm 0.52}$ & $11.75_{\pm 0.58}$ & $\mathbf{23.10}_{\pm 0.54}$ & $72.51_{\pm 0.23}$ & $\mathbf{42.27}_{\pm 0.70}$ & $102.84_{\pm 0.35}$ & $39.87_{\pm 3.53}$ & $20.17_{\pm 1.30}$\\
 \midrule
\multirow{2}{*}{Llama-2-7B} & PT & $26.52_{\pm 1.82}$ & $10.00_{\pm 0.96}$ & $19.97_{\pm 1.17}$ & $69.69_{\pm 0.80}$ & $41.18_{\pm 0.58}$ & $101.90_{\pm 1.08}$ & $32.45_{\pm 2.33}$ & $18.82_{\pm 2.16}$\\
 & LoRA & $26.70_{\pm 1.07}$ & $10.11_{\pm 0.50}$ & $20.53_{\pm 0.76}$ & $69.79_{\pm 0.54}$ & $41.18_{\pm 0.34}$ & $102.31_{\pm 0.52}$ & $40.01_{\pm 4.08}$ & $12.72_{\pm 1.28}$\\
 \midrule
\multirow{3}{*}{mBART} & FT & $24.07_{\pm 0.07}$ & $6.57_{\pm 0.01}$ & $16.41_{\pm 0.03}$ & $68.66_{\pm 0.00}$ & $35.57_{\pm 0.00}$ & $97.21_{\pm 0.00}$ & $56.10_{\pm 0.00}$ & $1.68_{\pm 0.00}$\\
 & PT & $29.22_{\pm 0.47}$ & $8.96_{\pm 0.80}$ & $20.46_{\pm 0.70}$ & $72.48_{\pm 0.31}$ & $41.01_{\pm 0.26}$ & $ 103.88_{\pm 1.29}$ & $56.95_{\pm 3.16}$ & $27.35_{\pm 4.86}$\\
 & LoRA & $29.60_{\pm 1.01}$ & $10.22_{\pm 0.79}$ & $21.44_{\pm 0.66}$ & $72.38_{\pm 0.96}$ & $41.18_{\pm 0.50}$ & $103.94_{\pm 1.35}$ & $\mathbf{61.34}_{\pm 1.77}$ & $19.40_{\pm 4.61}$\\
 \midrule
\multirow{3}{*}{mBARThez} & FT & $16.47_{\pm 0.01}$ & $5.28_{\pm 0.02}$ & $13.08_{\pm 0.05}$ & $65.96_{\pm 0.00}$ & $34.70_{\pm 0.00}$ & $96.95_{\pm 0.00}$ & $76.12_{\pm 0.00}$ & $11.02_{\pm 0.00}$\\
 & PT & $32.46_{\pm 0.74}$ & $11.36_{\pm 0.38}$ & $22.62_{\pm 0.60}$ & $73.57_{\pm 0.18}$ & $41.79_{\pm 0.77}$ & $104.17_{\pm 0.19}$ & $59.61_{\pm 1.52}$ & $20.26_{\pm 2.39}$\\
 & LoRA & $\mathbf{32.88}_{\pm 0.29}$ & $\mathbf{11.81}_{\pm 0.31}$ & $\mathbf{23.10}_{\pm 0.29}$ & $\mathbf{73.73}_{\pm 0.14}$ & $41.48_{\pm 0.34}$ & $\mathbf{104.21}_{\pm 0.20}$ & $56.52_{\pm 0.80}$ & $16.89_{\pm 1.40}$\\
 \bottomrule
\end{tabular}
}

\end{table*} 
\paragraph{Prefix-tuning} 
introduces a lightweight set of trainable vectors that are prepended to the key and value inputs of the Transformer multi-head attention mechanism~\cite{vaswani_attention_2017}. Formally, for each attention head \( i \), prefix-tuning prepends learned vectors \( P_K^i \in \mathbb{R}^{\rho \times \dhead} \) and \( P_V^i \in \mathbb{R}^{\rho \times \dhead} \), each of length \(\rho\), to the projected keys and values, respectively. The resulting attention computation for the \(i\)-th head is expressed as:

\begin{equation}
\text{head}_i = \text{Attention}\bigl(Q W_Q^i, [P_K^i; K W_K^i], [P_V^i; V W_V^i]\bigr)
\label{eq:prefix}
\end{equation}

\noindent where \( Q, K, V \in \mathbb{R}^{L \times \dmodel }\) denote the query, key, and value matrices derived from an input sequence of length \(L\), and
\( W_Q^i, W_K^i, W_V^i \in \mathbb{R}^{\dmodel \times \dhead} \) are the frozen projection matrices associated with the \(i\)-th attention head.
The notation \([;]\) denotes the concatenation function. To enhance the stability of prefix optimization, the number of trainable parameters is increased by employing a dedicated two-layer feed-forward network for re-parameterizing the prefix associated with each attention type. This network features an intermediate hidden dimension \(h_{\text{MLP}}\) and enables a richer parameterization of the prefix vectors. 

\paragraph{LoRA} offers an efficient alternative to full fine-tuning by introducing a low-rank decomposition of the linear model's weight matrices. Instead of updating the full weight matrix $W_0 \in \mathbb{R}^{d \times k}$, LoRA expresses it as a sum of the original weights and a trainable low-rank perturbation. Specifically, the update is represented by two smaller matrices: $B \in \mathbb{R}^{d \times r}$ and $A \in \mathbb{R}^{r \times k}$, with $r \ll \min(d, k)$. Here, $d$ and $k$ denote the input and output dimensions of the layer, respectively. The low-rank component is scaled by a factor $\alpha$ to control the magnitude of the update, ensuring minimal interference with the pre-trained backbone:

\begin{equation}
    \begin{split}
        h = W_0x + \frac{\alpha}{r}BAx
    \end{split}
    \label{eq:lora}
\end{equation}

\noindent
LoRA can be seamlessly integrated into Transformer architectures by applying it to each linear transformation, including the attention projection matrices $W_Q, W_K, W_V$, and $W_O$.

\subsection{Evaluation Metrics}

Since no dedicated evaluation metrics exist for ETR generation, we propose assessing it using standard summarization and text simplification metrics. For summarization, we report F1-scores for ROUGE-1, ROUGE-2, ROUGE-L \cite{lin_rouge_2004}, and BERTScore \cite{zhang_bertscore_2020}. For simplification, we include SARI \cite{xu_optimizing_2016}, Kandel-Moles Readability Estimate (KMRE) \cite{kandel_application_1958}, and novelty ratio for unigrams \cite{kamal_eddine_barthez_2021}. BLEU is excluded, as it is unsuitable for text simplification \cite{xu_optimizing_2016, sulem_simple_2018}. 

\subsection{Experimental Setup}\label{subsec:experiment}

All PLMs are trained for 30 epochs, while LLMs are trained for 5 epochs, using the AdamW optimizer~\cite{loshchilov_decoupled_2019} with the following parameters: $\epsilon=10^{-9}$, $\beta_1=0.9$, $\beta_2=0.999$, and a weight decay of $\lambda=0.01$. A linear learning rate scheduler with a 10\% warm-up ratio is employed. The training batch size is fixed at 8, with no gradient accumulation. The learning rate is chosen from the set $\{1\cdot10^{-5}, 2\cdot10^{-5}, 5\cdot10^{-5}, 1\cdot10^{-4}\}$, and hyperparameter tuning for prefix-tuning and LoRA is performed to maximize the harmonic mean of SARI, ROUGE-L, and BERTScore. Each best model is selected following a hyperparameter search policy using grid search. 

In particular for prefix-tuning, we explore prefix length $\rho \in \{10, 50, 150, 250, 500\}$ and re-parametrization MLP hidden size $h_{\text{MLP}} \in \{256, 512, 1024, 2048\}$. 

For LoRA, we explore $r \in \{8, 16, 32, 64, 128\}$, $dropout \in \{0.0, 0.05, 0.1\}$, and which matrices to adapt for the self-attention and cross-attention layers $attn\_matrices \in \{W_Q, W_K, W_V, W_O, W_{QK}, W_{QV}, W_{KV}, W_{QKVO}\}$. To keep a 1:1 ratio so as not to overpower the backbone, we choose $\alpha = r$ \cite{lee_platypus_2023}. 

For evaluation, generation performance results are averaged over five runs, distinguishing our approach from most text generation studies that typically report results from a single run or fixed seed \cite{li_prefix-tuning_2021,martinez_exploring_2024}. The expert-centric model is the only one evaluated in a zero-shot setting.
\begin{table*}[t!]
\caption{
Performance metrics for fine-tuned models on ETR-fr, tested on the ETR-fr-politic test set. Results are reported as average with standard deviation over 5 runs. The best scores are highlighted in bold.
}\label{tab:politic_results}
\centering
\footnotesize
\resizebox{\textwidth}{!}{
\renewcommand{\arraystretch}{1.4}
\begin{tabular}{llcccccccc}
\toprule
 &  & { \textbf{ROUGE-1}} & { \textbf{ROUGE-2}} & { \textbf{ROUGE-L}} & { \textbf{BERT-$F_1$}} & { \textbf{SARI}} & { \textbf{KMRE}} & { \textbf{Comp. ratio}} & { \textbf{Novelty}}\\
 \midrule
\multirow{2}{*}{Mistral-7B} & PT & $22.56_{\pm 11.68}$ & $7.92_{\pm 4.56}$ & $16.95_{\pm 8.45}$ & $63.29_{\pm 9.14}$ & $36.71_{\pm 1.22}$ & $80.34_{\pm 3.89}$ & $-9.53_{\pm 18.26}$ & $12.77_{\pm 9.08}$\\
 & LoRA & $33.16_{\pm 1.34}$ & $12.04_{\pm 0.84}$ & $25.00_{\pm 0.92}$ & $69.45_{\pm 0.53}$ & $39.39_{\pm 0.40}$ & $79.66_{\pm 0.39}$ & $7.90_{\pm 4.60}$ & $15.33_{\pm 1.98}$\\
 \midrule
\multirow{2}{*}{Llama-2-7B} & PT & $24.64_{\pm 3.04}$ & $8.90_{\pm 1.42}$ & $19.44_{\pm 2.03}$ & $65.35_{\pm 1.46}$ & $37.74_{\pm 2.17}$ & $81.89_{\pm 1.01}$ & $-20.17_{\pm 19.57}$ & $22.54_{\pm 3.44}$\\
 & LoRA & $27.79_{\pm 0.75}$ & $11.03_{\pm 0.18}$ & $21.24_{\pm 0.35}$ & $66.83_{\pm 0.37}$ & $39.14_{\pm 0.15}$ & $73.49_{\pm 0.98}$ & $-9.22_{\pm 4.44}$ & $15.41_{\pm 0.94}$\\
 \midrule
\multirow{2}{*}{mBART} & PT & $28.58_{\pm 0.79}$ & $9.72_{\pm 1.42}$ & $21.20_{\pm 1.60}$ & $67.94_{\pm 0.49}$ & $\mathbf{40.42}_{\pm 0.77}$ & $\mathbf{86.98}_{\pm 1.73}$ & $46.24_{\pm 3.13}$ & $39.03_{\pm 6.68}$\\
 & LoRA & $31.72_{\pm 1.57}$ & $10.61_{\pm 1.05}$ & $24.07_{\pm 0.95}$ & $69.05_{\pm 1.25}$ & $39.78_{\pm 0.81}$ & $85.82_{\pm 1.61}$ & $41.92_{\pm 2.06}$ & $\mathbf{34.31}_{\pm 2.34}$\\
 \midrule
\multirow{2}{*}{mBARThez} & PT & $36.79_{\pm 0.68}$ & $14.43_{\pm 0.72}$ & $26.95_{\pm 0.65}$ & $71.11_{\pm 0.35}$ & $39.23_{\pm 0.60}$ & $81.92_{\pm 0.80}$ & $37.86_{\pm 2.43}$ & $12.58_{\pm 3.57}$\\
 & LoRA & $\mathbf{38.12}_{\pm 0.32}$ & $\mathbf{14.73}_{\pm 0.67}$ & $\mathbf{28.11}_{\pm 0.40}$ & $\mathbf{71.31}_{\pm 0.32}$ & $40.35_{\pm 0.37}$ & $81.58_{\pm 0.50}$ & $\mathbf{35.37}_{\pm 1.30}$ & $16.74_{\pm 2.20}$\\
 \bottomrule
\end{tabular}
}

\end{table*}
\section{Quantitative and Qualitative Results}\label{part:results}

To rigorously evaluate the various ETR generation models, we propose a dual approach: a quantitative evaluation using both in-domain and out-of-domain test sets, and a qualitative assessment through manual evaluation by linguist-experts, based on 36 questions from the European ETR guidelines.

\subsection{In-Domain Quantitative Results}\label{part:in_domain_quantitative_results}

Table~\ref{tab:results} presents the evaluation metrics for all ETR generation models on the ETR-fr test set. 
In the expert-centric pipelines, MUSS achieves the best ROUGE-1 (28.11) and ROUGE-2 (8.87) but shows low compression (6.62) and novelty (15.00), indicating a conservative style. BARThez performs moderately (ROUGE-1: 22.85). The combined pipelines trade fidelity for abstraction: MUSS+BARThez yields the highest compression (75.62), best SARI (37.56), and greatest novelty (37.48), though with weaker ROUGE (20.15/5.36/13.58) and BERTScore (66.85).

For fine-tuned models, PEFT methods outperform full fine-tuning, aligning with the findings of \cite{vu_overcoming_2022}. Mistral-7B with LoRA achieves strong results, with ROUGE-L (23.10), SARI (42.27), and novelty (20.17). Llama-2-7B, in both prefix-tuning and LoRA configurations, delivers competitive performance, with ROUGE-L scores of 19.97 and 20.53, respectively. 

Among the fine-tuned models, mBART with LoRA exhibits the best compression ratio (61.34) (closest to the test split reference), while maintaining strong ROUGE-1 (29.60) and ROUGE-2 (10.22) scores. 
The PLM mBARThez with LoRA achieves the best overall performance, with the highest ROUGE-1 (32.88), ROUGE-2 (11.81), ROUGE-L (23.10), BERTScore (73.73), and KMRE (104.21). Interestingly, prefix-tuning delivers results comparable to LoRA across both PLMs and LLMs.

\subsection{Out-of-Domain Quantitative Results}

Table \ref{tab:politic_results} illustrates the performance of fine-tuned models on ETR-fr when evaluated on ETR-fr-politic test set. Similarly to results in \S\ref{part:in_domain_quantitative_results}, mBARThez achieves the highest scores across most metrics, particularly with the LoRA configuration. It records the top ROUGE-1 (38.12), ROUGE-2 (14.73), and ROUGE-L (28.11), along with the highest BERTScore (71.31) and a strong SARI score (40.35). Overall, LoRA emerges as the superior fine-tuning strategy, consistently yielding higher performance across all models compared to prefix-tuning. Additionally, the lower standard deviations associated with LoRA, especially for Mistral-6B and mBARThez, underline their stability. However, the analysis reveals that LLMs exhibit a negative compression rate, indicating challenges in replicating summarization behavior effectively.

\subsection{Manual Qualitative Results}\label{part:results_results}


\begin{figure}[!ht]
    \centering
    \begin{tikzpicture}
        \node[inner sep=0pt] (img1) at (0,0) {\includegraphics[width=.9\linewidth]{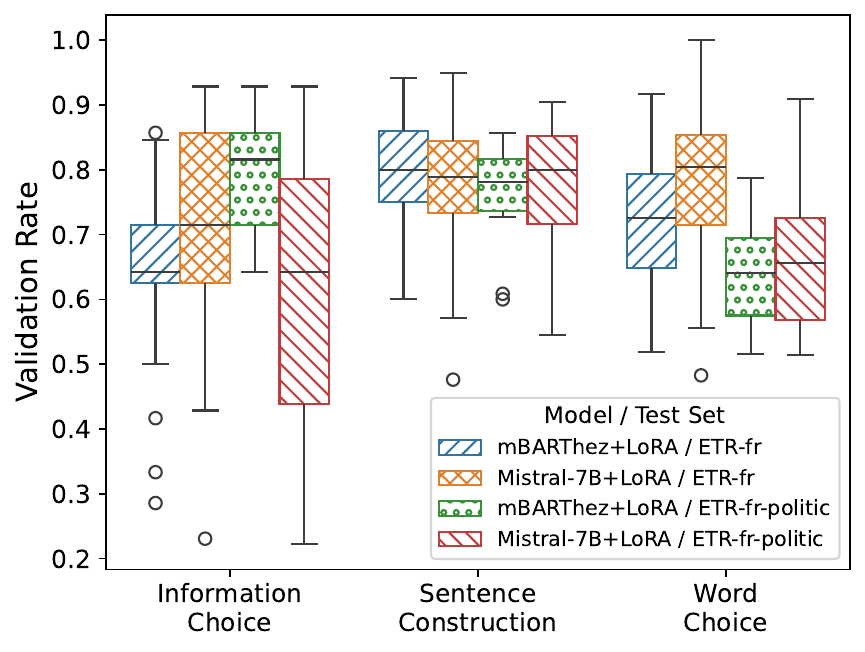}};
        \node[anchor=south] at ([yshift=-15pt]img1.south) {(a) Validation Rate of ETR Criteria};
        \label{fig:boxes_manual_etr_eval}
    \end{tikzpicture}


    \begin{tikzpicture}
        \node[inner sep=0pt] (img2) at (0,0) {\includegraphics[width=.9\linewidth]{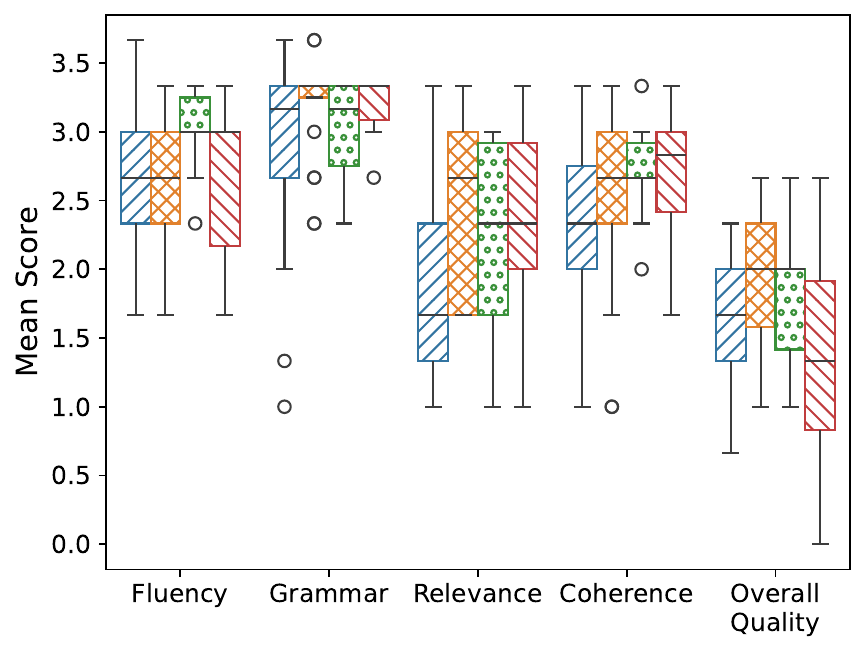}};
        \node[anchor=south] at ([yshift=-15pt]img2.south) {(b) Quality Score of Generation Quality Criteria};
        \label{fig:boxes_manual_quality}
    \end{tikzpicture}


    \caption{Manual evaluation comparisons. \textbf{(a)} Assessments from 28 ETR guidelines questions grouped into three categories. \textbf{(b)} Assessments from 8 text generation questions grouped into five categories.}
    \label{fig:boxes_manual}
\end{figure}

Manual evaluation is essential for assessing the quality of ETR text production and compliance with European ETR guidelines. These guidelines consist of 57 questions categorized by topic and weighted by importance, forming a comprehensive framework for evaluating clarity, simplicity, and accessibility. By following these standards, the evaluation process ensures linguistic accuracy while also verifying that the texts meet cognitive requirements, making them understandable, engaging, and suitable for the target audience.

To validate our approach, we conduct a human evaluation with three linguist-experts\footnote{The linguist-experts, all second-year Master’s students in Language Studies, received dedicated training sessions to prepare for the evaluation task. Their participation was voluntary and uncompensated, and they were kept unaware of the model development to ensure unbiased assessments.} across the ETR-fr and ETR-fr-politic test sets. The assessment begins by focusing on the most critical criteria from the ETR guidelines checklist, including Information Choices (IC), Sentence Construction (SC), Word Choice (WC), and Illustrations\footnote{Results for Illustrations are not presented, as this criterion did not apply to most of the evaluated texts.}, and consisting of 29 individual questions. Additionally, we evaluate general criteria commonly used in automatic text generation, such as Fluency, Grammar/Spelling, Relevance, Textual Coherence, and Overall Perceived Quality, gathered in an additional 8 individual questions. ETR criteria are assessed using a binary scale (respected, not respected), while human judgments are rated on a 5-point Likert scale (0–4).

For each model, annotators were assigned to evaluate 20 texts from ETR-fr and 10 from ETR-fr-politic, randomly sampled. All annotators assessed the same set of texts, ensuring consistency in the evaluation process across models and datasets. 
The averaged inter-annotator agreement over the 36 criteria is $\alpha = 0.07$\footnote{It reaches $0.20$ for a binarized aggregated scores.}~\cite{Krippendorff2004-ix}. 

Figure \ref{fig:boxes_manual} (a) presents the results of the ETR guidelines-based evaluation for the two best competing models: mBARThez+LoRA and Mistral-7B+LoRA. Unlike the automatic evaluation, the manual assessment shows that Mistral-7B+LoRA achieves the highest scores for IC and WC, while mBARThez+LoRA excels in SC on the ETR-fr test set. Interestingly, the trend is almost reversed on ETR-fr-politic, where mBARThez+LoRA scores highest for IC and performs comparably to Mistral-7B+LoRA for WC and SC. Additionally, for both test sets, the mBARThez model exhibits the lowest dispersion score, indicating greater stability in generation.

Figure \ref{fig:boxes_manual} (b) presents the manual evaluation results for text generation quality and accuracy. Similar to the ETR-based assessment, Mistral-7B+LoRA achieves the highest scores for most criteria on the ETR-fr test set, though mBARThez+LoRA performs equally well in Fluency. However, the trend shifts significantly in the out-of-domain setting, where mBARThez+LoRA emerges as the top-performing model for Overall Perceived Quality and Fluency.

In summary, Mistral-7B+LoRA appears to overfit on ETR-fr, while mBARThez+LoRA demonstrates better generalization for ETR generation, achieving the highest results on ETR-fr-politic while maintaining strong performance on ETR-fr.
\section{Limitations and Perspectives}
\label{sec:perspectives}

The automatic evaluation of text generation models remains an open issue~\cite{jamet_difficulty_2024}. We argue that specific metrics should be developed for ETR generation, considering aspects such as novelty ratio, repetition, and coherence. Indeed, evaluation metrics for summarization and text simplification do not capture all characteristics of ETR generation, even when combined into a unique score as used in this work. 

The low inter-annotator agreement observed in \S\ref{part:results_results} may be explained by the high number of ETR criteria (>30), which is known to reduce agreement levels~\cite{bayerl-paul-2011-determines}, as well as the abstract nature of these criteria~\cite{Canut2020-gi}, which introduces subjectivity. Improved formalization or targeted annotator training, especially with disabled users, could help mitigate this variability.

While our dataset is limited in size, cross-lingual transfer remains particularly challenging due to the lack of data in other languages, especially in English. Additionally, \citet{martinez_exploring_2024} demonstrate that the translate-simplify-retranslate strategy is ineffective for ETR, often resulting in incorrect outputs. Using data from other languages also necessitates a rigorous, manual translation process involving native speakers to ensure accessibility, which restricts scalability. Although developing a multilingual model could alleviate this issue, it would still require a large-scale protocol for manual ETR transcription to create reliable resources in English.

Reinforcement learning from human feedback (RLHF)~\cite{ouyang_training_2022} could further refine ETR generation by aligning model outputs with user preferences. Collecting high-quality preference data from both expert writers and cognitively disabled users is essential to train reward models that guide optimization of language models. 
This process would involve annotation tasks where users rank generated texts by clarity, accessibility, and engagement.
Expanding RLHF data collection across languages and cognitive conditions would ensure that models generate texts that are both contextually appropriate and widely usable. Moreover, this process could be a step toward automating the acquisition of the European ETR label.

\section{Conclusion}
\label{sec:conclusion}

This paper addresses ETR text generation for individuals with cognitive impairments, aiming to enhance their self-determination and autonomy by bridging the digital divide. To support this objective, we introduced the ETR-fr dataset, a set of 523 pairs of ETR-aligned texts, and conducted an extensive empirical study using multilingual PLMs and LLMs. Our findings show that ETR generation differs significantly from traditional text simplification and summarization tasks, requiring a focused approach on cognitive accessibility. Remarkably, the small mBARThez model, combined with LoRA tuning, performs on par with larger LLMs, achieving the best results in ROUGE and BERTScore, as well as highly competitive indicators for simplification assessment, across both in-domain and out-of-domain settings. The manual evaluation conducted by three linguist-experts also highlights that the LLM-based approach tends to overfit to the main task, whereas the lightweight approach generalizes better, achieving the highest results on the political test set while maintaining strong performance on the original task.
\begin{ack}
We are grateful to François Baudez Publishing for generously providing free access to the \falc book collection to build our dataset.

Finally, the present work was performed using computing resources of CRIANN (Normandy, France).
\end{ack}

\bibliography{references}

\end{document}